\documentclass[11pt]{article}
\usepackage[margin=1in]{geometry}
\usepackage[utf8]{inputenc}
\usepackage[T1]{fontenc}
\usepackage{hyperref}
\usepackage{url}
\usepackage{booktabs}
\usepackage{amsfonts}
\usepackage{amsmath}
\usepackage{microtype}
\usepackage{graphicx}
\usepackage{float}
\usepackage{natbib}
\usepackage{authblk}
\usepackage{tikz}
\usetikzlibrary{positioning, arrows.meta, fit, backgrounds}

\title{Adapting VACE for Real-Time Autoregressive Video Diffusion}

\author[1]{Ryan Fosdick\thanks{ryan@livepeer.org}}
\affil[1]{Daydream}

\date{}

\begin{document}

\maketitle

\begin{abstract}
We describe an adaptation of VACE (Video All-in-one Creation and Editing) for real-time autoregressive video generation. VACE provides unified video control (reference guidance, structural conditioning, inpainting, and temporal extension) but assumes bidirectional attention over full sequences, making it incompatible with streaming pipelines that require fixed chunk sizes and causal attention. The key modification moves reference frames from the diffusion latent space into a parallel conditioning pathway, preserving the fixed chunk sizes and KV caching that autoregressive models require. This adaptation reuses existing pretrained VACE weights without additional training. Across 1.3B and 14B model scales, VACE adds 20--30\% latency overhead for structural control and inpainting, with negligible VRAM cost relative to the base model. Reference-to-video fidelity is severely degraded compared to batch VACE due to causal attention constraints. A reference implementation is available at \url{https://github.com/daydreamlive/scope}.
\end{abstract}

\section{Introduction}

Real-time video generation models such as LongLive~\citep{yang2026longlive}, Krea Realtime Video~\citep{millon2025krea}, and StreamDiffusion V2~\citep{feng2025streamdiffusionv2} generate video autoregressively in chunks using causal attention. Each chunk attends only to itself and past frames, enabling KV caching and bounded memory usage. However, these models lack the control capabilities available to batch video generation models: reference guidance, structural conditioning, and selective editing. Building these capabilities from scratch would require extensive retraining.

VACE~\citep{jiang2025vace} provides unified video control for batch-oriented diffusion models, supporting reference-to-video generation, video-to-video structural control, masked editing (inpainting/outpainting), temporal extension, and arbitrary compositions of these capabilities. However, VACE assumes bidirectional attention and processes full video sequences at once, making it incompatible with streaming generation.

This paper presents an adaptation that enables VACE's control capabilities in streaming autoregressive pipelines. We validate on five Wan2.1-based autoregressive pipelines spanning 1.3B and 14B parameter scales. Our contributions are:

\begin{itemize}
    \item An architectural modification that moves reference frames from the diffusion latent space into a parallel conditioning pathway, resolving the incompatibility between VACE's reference handling and fixed-size chunk processing.
    \item Demonstration that pretrained VACE Context Block weights transfer directly to the adapted architecture without fine-tuning, due to the preservation of zero-initialized hint injection structure.
    \item Empirical validation that structural control (depth, scribble, optical flow, layout), masked generation (inpainting, outpainting), and temporal extension all function at real-time rates (17--22 FPS) with the 1.3B model on consumer hardware in streaming contexts.
    \item Empirical validation across Wan2.1 1.3B and 14B model scales using the same adaptation code without per-model modifications.
\end{itemize}

The adaptation enables pretrained VACE Context Blocks to function in causal chunked pipelines without retraining or modifying base model weights.

\section{Background}

\subsection{Autoregressive Video Generation}

Autoregressive video diffusion models~\citep{yang2026longlive, ji2025memflow, lu2025rewardforcing} generate video in fixed-size chunks using causal attention patterns. Each chunk of latent frames is denoised conditioned on cached key-value pairs from previously generated chunks. This design enables streaming generation with bounded memory, but constrains the model to causal (past-only) attention.

\subsection{VACE: Unified Video Control}

VACE~\citep{jiang2025vace} unifies video control through three optional conditioning inputs combined with a text prompt:

\begin{itemize}
    \item \textbf{Source video} (\texttt{src\_video}): Conditioning signal such as depth maps, pose skeletons, or video to edit.
    \item \textbf{Source mask} (\texttt{src\_mask}): Defines reactive (white, to be generated) vs.\ inactive (black, to be preserved) regions.
    \item \textbf{Reference images} (\texttt{src\_ref\_images}): Style or subject guidance images.
\end{itemize}

VACE processes these inputs through a parallel set of transformer blocks (Context Blocks) that produce ``hints,'' additive signals injected into the main Diffusion Transformer (DiT)~\citep{peebles2023dit} pathway via zero-initialized linear projections. The base DiT is frozen; only the Context Blocks are trained (Context Adapter Tuning).

The publicly released VACE weights target the Wan video generation model~\citep{wang2025wan}, which provides 1.3B and 14B parameter variants with a 3D VAE for temporal compression.

\section{The Architectural Problem}

\subsection{Reference Handling in Original VACE}

VACE concatenates reference frames directly into the diffusion latent sequence alongside the video latents. The model processes this combined sequence with bidirectional attention, then strips reference frames from the output after denoising.

This approach has three incompatibilities with streaming generation:

\begin{enumerate}
    \item \textbf{Variable sequence lengths.} Different tasks require different numbers of reference frames, preventing the fixed-size chunk processing that streaming models require.
    \item \textbf{KV cache contamination.} Concatenated references become part of the model's causal history, cached and attended to as if they were previously generated frames. This is semantically incorrect: references should guide generation, not be treated as historical context. Furthermore, RoPE~\citep{su2021roformer} positional encodings are baked into cached K/V tensors, so removing references would require recomputing the entire cache.
    \item \textbf{Post-processing overhead.} Reference frames must be identified and removed after each denoising step.
\end{enumerate}

\subsection{Adaptation: Separate Conditioning Pathway}
\label{sec:adaptation}

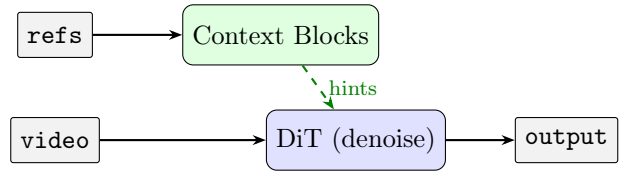
\begin{figure}[h]
\centering
\begin{tikzpicture}[
    box/.style={draw, rounded corners, minimum height=0.8cm, minimum width=1.8cm, font=\small, align=center},
    data/.style={draw, rounded corners=1pt, minimum height=0.6cm, font=\small\ttfamily, fill=gray!10},
    arr/.style={-{Stealth[length=5pt]}, thick},
    darr/.style={-{Stealth[length=5pt]}, thick, dashed, color=green!50!black},
]
\node[font=\small\bfseries] at (0, 0) (orig-label) {Original VACE};

\node[data] at (0, -1) (refs1) {refs};
\node[data] at (1.2, -1) (vid1) {video};
\node[draw, thick, rounded corners, fit=(refs1)(vid1), inner sep=3pt] (concat) {};

\node[box, fill=blue!12] at (4, -1) (denoise1) {DiT\\(denoise)};
\node[box, fill=orange!15] at (6.5, -1) (strip) {Strip refs};
\node[data] at (8.8, -1) (out1) {output};

\draw[arr] (concat) -- (denoise1);
\draw[arr] (denoise1) -- (strip);
\draw[arr] (strip) -- (out1);

\node[font=\small\bfseries] at (0, -2.8) (adapt-label) {Streaming Adaptation};

\node[data] at (0, -3.8) (refs2) {refs};
\node[box, fill=green!12] at (3, -3.8) (ctx) {Context Blocks};

\node[data] at (0, -5.2) (vid2) {video};
\node[box, fill=blue!12] at (4, -5.2) (denoise2) {DiT (denoise)};
\node[data] at (6.8, -5.2) (out2) {output};

\draw[arr] (refs2) -- (ctx);
\draw[arr] (vid2) -- (denoise2);
\draw[darr] (ctx) -- node[right, font=\scriptsize, color=green!50!black] {hints} (denoise2);
\draw[arr] (denoise2) -- (out2);

\end{tikzpicture}
\caption{Original VACE concatenates references into the latent sequence, requiring post-hoc stripping. The streaming adaptation processes references through separate Context Blocks that inject hints into the DiT pathway, preserving fixed chunk sizes.}
\label{fig:arch}
\end{figure}

We move reference frames out of the diffusion latent space into a parallel conditioning pathway (Figure~\ref{fig:arch}). Video latents are denoised alone while reference frames are processed separately by Context Blocks that produce hints: additive signals injected into the main DiT pathway via zero-initialized projections scaled by a user-controllable context scale. This preserves fixed chunk sizes regardless of how many references are provided.

\section{Why Pretrained Weights Transfer}

The publicly released VACE weights use Context Adapter Tuning: the base DiT is frozen, and separate Context Blocks are trained to process conditioning inputs and inject hints. The Context Blocks are already trained to:

\begin{itemize}
    \item Encode reference information into a representation suitable for hint generation
    \item Generate hints that modulate the main DiT pathway
    \item Apply zero-initialized projections for controlled influence
\end{itemize}

\begin{table}[h]
\centering
\caption{Comparison of original VACE and the streaming adaptation.}
\label{tab:comparison}
\begin{tabular}{lll}
\toprule
\textbf{Component} & \textbf{Original VACE} & \textbf{Streaming Adaptation} \\
\midrule
Reference input & Concatenated into noisy latents & Separate conditioning tensor \\
Context Block inputs & Full sequence (refs + video) & References only \\
Hint injection target & Mixed ref+video sequence & Video-only sequence \\
Attention pattern & Bidirectional & Causal \\
\bottomrule
\end{tabular}
\end{table}

The Context Blocks themselves are unchanged; they process references and produce hints using the same weights. The adaptation changes \emph{where} references enter the pipeline and \emph{where} hints are injected. For structural control and masking, where hints are spatially local, control adherence is preserved (Table~\ref{tab:quality}); for reference-to-video, where the blocks relied on cross-attention between references and video latents, quality is severely degraded (Section~\ref{sec:limitations}).

Concretely, each DiT block $i$ with a corresponding Context Block receives a hint via residual addition:
\begin{equation}
    \mathbf{x}_i = \text{DiTBlock}_i(\mathbf{x}_{i-1}) + \alpha \cdot W_{\text{proj}}^{(i)} \, \mathbf{h}_i
\end{equation}
where $\mathbf{h}_i$ is the hint from Context Block $i$, $W_{\text{proj}}^{(i)}$ is a zero-initialized linear projection, and $\alpha$ is the user-controllable context scale. Since $W_{\text{proj}}$ is zero at initialization, hints have no effect until trained, and the trained weights remain valid in the adapted architecture because neither the projection nor the residual structure changes.

\section{Streaming Compatibility and Capabilities}

Beyond reference handling (Section~\ref{sec:adaptation}), most VACE primitives transfer to streaming contexts with only cache management changes. Masks, control signals (depth, pose, optical flow, scribble, grayscale, layout), dual-stream encoding, and hint injection all function with the same core mechanisms.

\begin{table}[h]
\centering
\caption{Streaming compatibility of VACE components.}
\label{tab:compat}
\small
\begin{tabular}{llp{5.5cm}}
\toprule
\textbf{Component} & \textbf{Compatibility} & \textbf{Notes} \\
\midrule
Masks & Full & Separate temporal autoencoder caches for inactive/reactive streams \\
Control signals (depth, pose) & Full & Per-chunk processing, same encoding \\
Dual-stream encoding & Full & Cache separation prevents contamination \\
Hint injection & Full & Residual addition unchanged \\
Reference images & Requires adaptation & Architectural change (Section~\ref{sec:adaptation}) \\
Temporal extension & Partial & Fixed chunk sizes limit first+last frame utility \\
\bottomrule
\end{tabular}
\end{table}

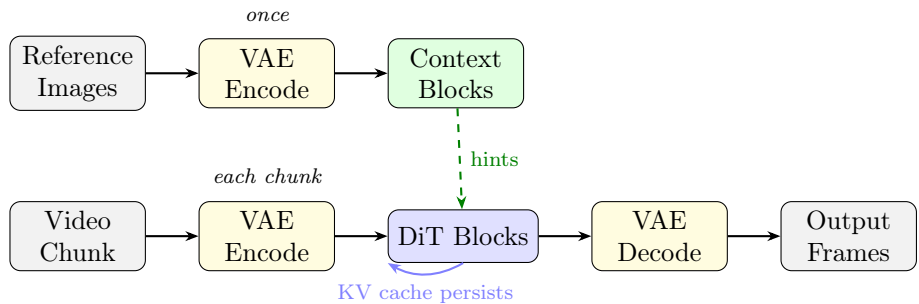
\begin{figure}[h]
\centering
\begin{tikzpicture}[
    node distance=0.5cm and 0.7cm,
    box/.style={draw, rounded corners, minimum height=0.7cm, minimum width=1.8cm, font=\small, align=center},
    arr/.style={-{Stealth[length=5pt]}, thick},
    darr/.style={-{Stealth[length=5pt]}, thick, dashed, color=green!50!black},
]
\node[box, fill=gray!10] (ref) {Reference\\Images};
\node[box, right=of ref, fill=yellow!15] (vae-ref) {VAE\\Encode};
\node[box, right=of vae-ref, fill=green!12] (ctx) {Context\\Blocks};

\draw[arr] (ref) -- (vae-ref);
\draw[arr] (vae-ref) -- (ctx);

\node[font=\scriptsize\itshape, above=0.1cm of vae-ref] {once};

\node[box, below=1.2cm of ref, fill=gray!10] (chunk) {Video\\Chunk};
\node[box, right=of chunk, fill=yellow!15] (vae-vid) {VAE\\Encode};
\node[box, right=of vae-vid, fill=blue!12] (dit) {DiT Blocks};
\node[box, right=of dit, fill=yellow!15] (dec) {VAE\\Decode};
\node[box, right=of dec, fill=gray!10] (out) {Output\\Frames};

\draw[arr] (chunk) -- (vae-vid);
\draw[arr] (vae-vid) -- (dit);
\draw[arr] (dit) -- (dec);
\draw[arr] (dec) -- (out);

\node[font=\scriptsize\itshape, above=0.1cm of vae-vid] {each chunk};

\draw[darr] (ctx) -- node[right, font=\scriptsize, color=green!50!black] {hints} (dit);

\draw[arr, color=blue!50, bend left=30] (dit.south) to node[below, font=\scriptsize, color=blue!50] {KV cache persists} (dit.south west);

\end{tikzpicture}
\caption{Per-chunk processing in the streaming VACE adaptation. Reference images are encoded once by Context Blocks; hints are injected into DiT blocks for each video chunk. The KV cache persists across chunks for autoregressive continuity.}
\label{fig:pipeline}
\end{figure}

\begin{figure}[H]
\centering
\setlength{\tabcolsep}{1pt}
\renewcommand{\arraystretch}{0.5}
\begin{tabular}{cccc}
& \small Input & \small Conditioning & \small Output \\[2pt]
\rotatebox{90}{\small ~~Depth} &
\includegraphics[width=0.145\textwidth]{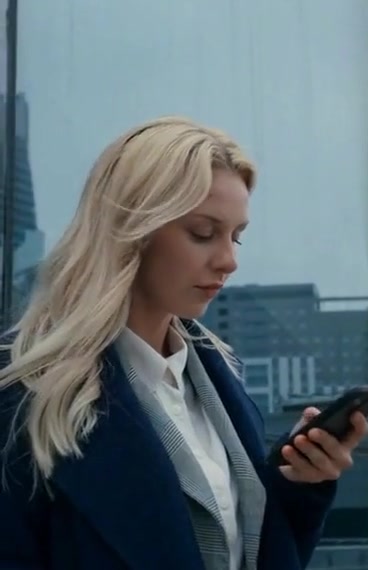} &
\includegraphics[width=0.145\textwidth]{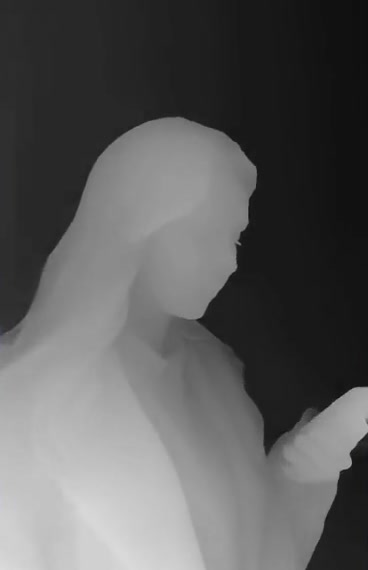} &
\includegraphics[width=0.145\textwidth]{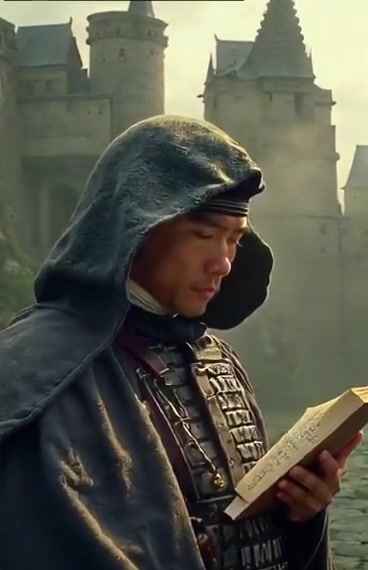} \\[1pt]
\rotatebox{90}{\small ~~Scribble} &
\includegraphics[width=0.145\textwidth]{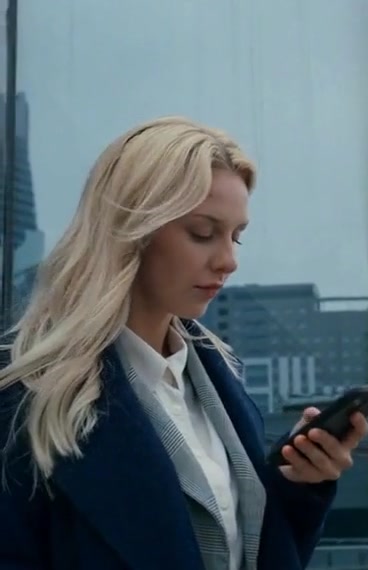} &
\includegraphics[width=0.145\textwidth]{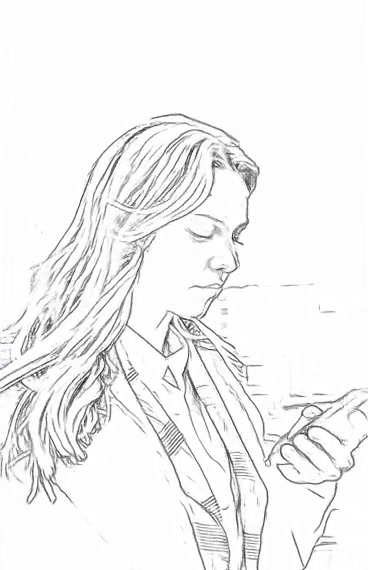} &
\includegraphics[width=0.145\textwidth]{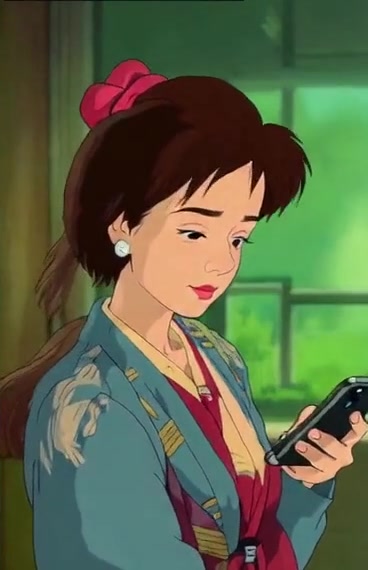} \\[1pt]
\rotatebox{90}{\small ~~Flow} &
\includegraphics[width=0.145\textwidth]{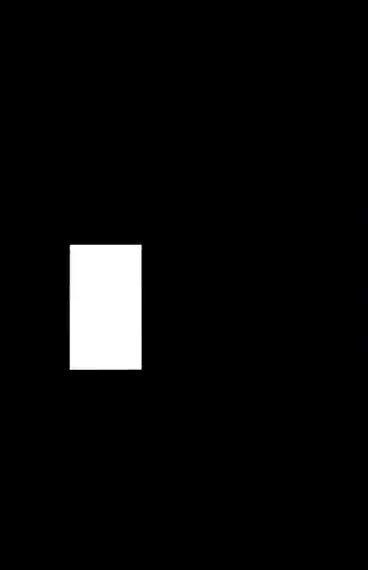} &
\includegraphics[width=0.145\textwidth]{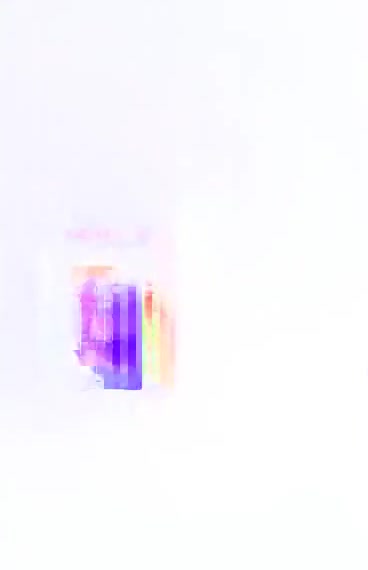} &
\includegraphics[width=0.145\textwidth]{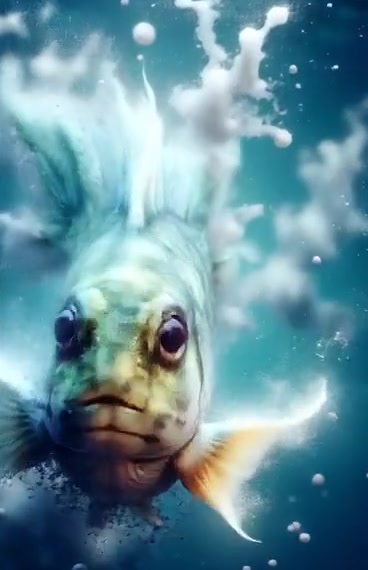} \\[1pt]
\rotatebox{90}{\small ~~Color} &
\includegraphics[width=0.145\textwidth]{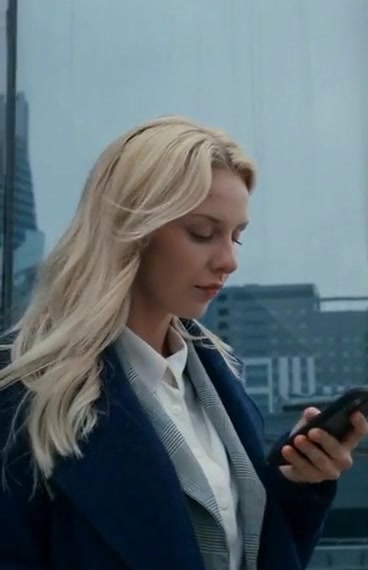} &
\includegraphics[width=0.145\textwidth]{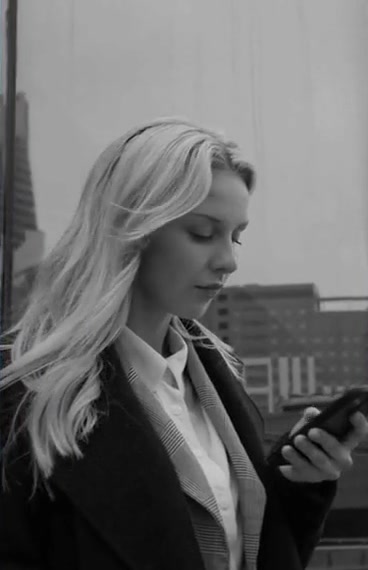} &
\includegraphics[width=0.145\textwidth]{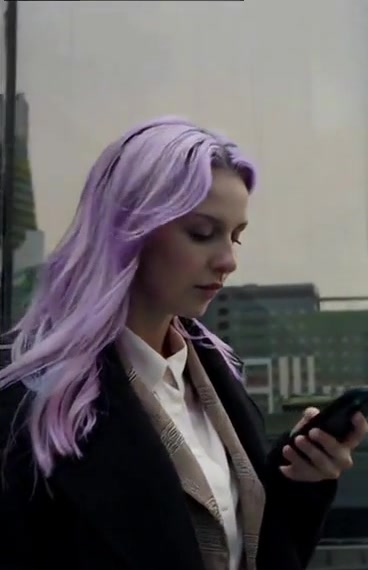} \\
\end{tabular}
\caption{Structural control modes. Each row: input frame, extracted conditioning signal, and generated output. Depth, scribble/edge, optical flow, and colorization (grayscale) controls shown.}
\label{fig:structural}
\end{figure}

\begin{figure}[H]
\centering
\setlength{\tabcolsep}{1pt}
\renewcommand{\arraystretch}{0.5}
\begin{tabular}{cccc}
& \small Input & \small Mask/Control & \small Output \\[2pt]
\rotatebox{90}{\small ~~Inpaint} &
\includegraphics[width=0.15\textwidth]{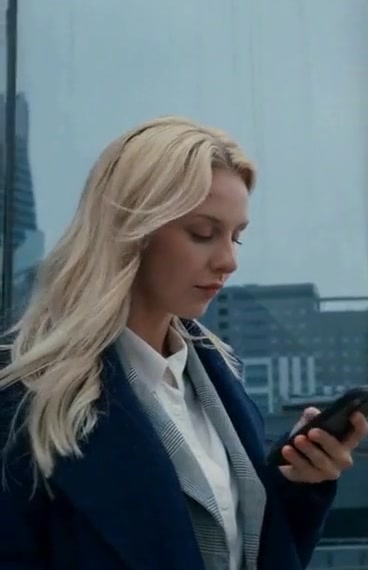} &
\includegraphics[width=0.15\textwidth]{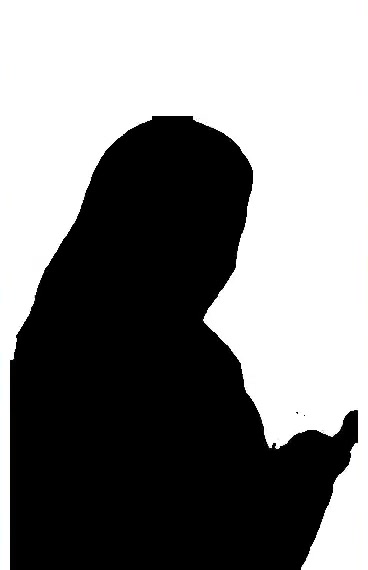} &
\includegraphics[width=0.15\textwidth]{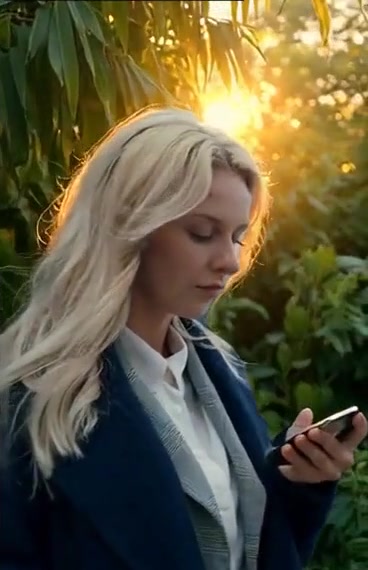} \\[1pt]
\rotatebox{90}{\small ~~LoRA} &
\includegraphics[width=0.15\textwidth]{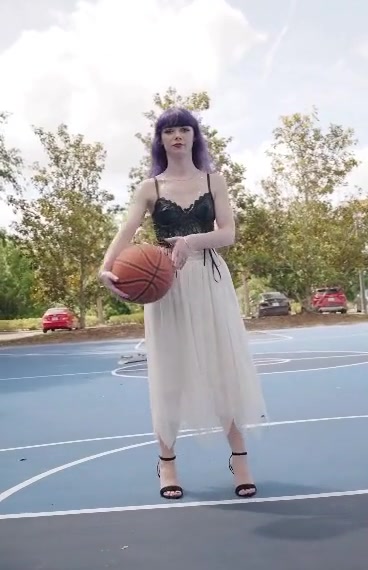} &
\includegraphics[width=0.15\textwidth]{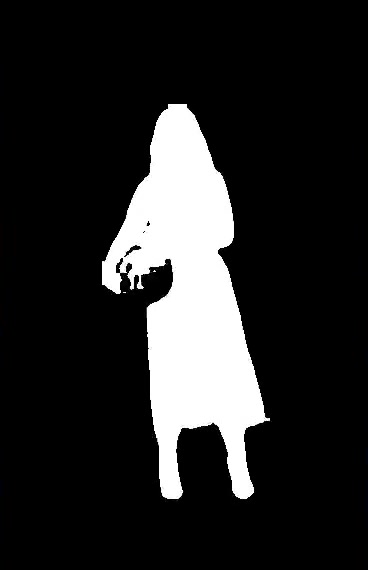} &
\includegraphics[width=0.15\textwidth]{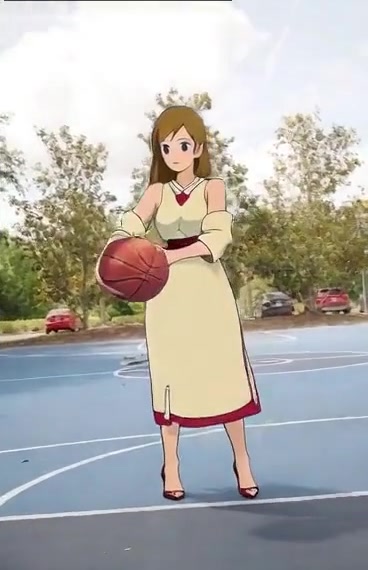} \\[1pt]
\rotatebox{90}{\small ~~Layout} &
\includegraphics[width=0.15\textwidth]{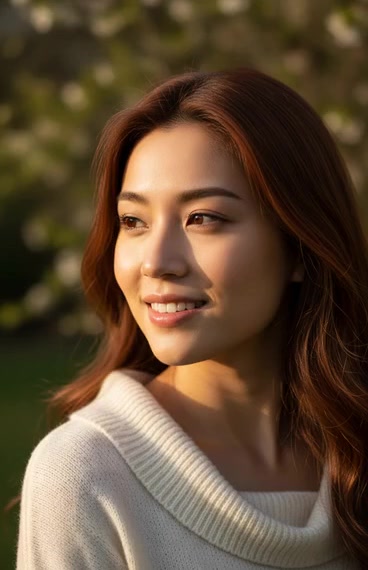} &
\includegraphics[width=0.15\textwidth]{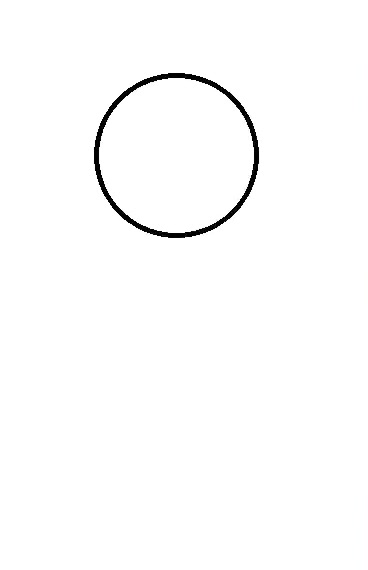} &
\includegraphics[width=0.15\textwidth]{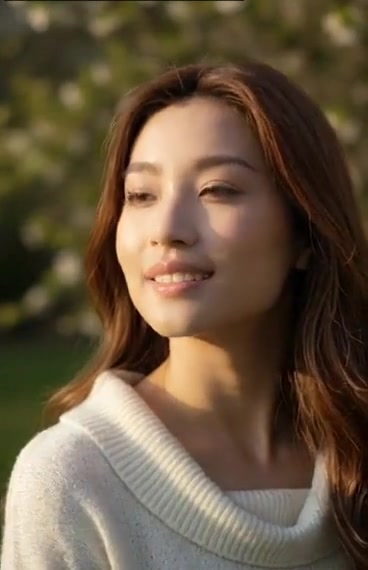} \\[1pt]
& \small Input & & \small Output \\[2pt]
\rotatebox{90}{\small ~~I2V} &
\includegraphics[width=0.15\textwidth]{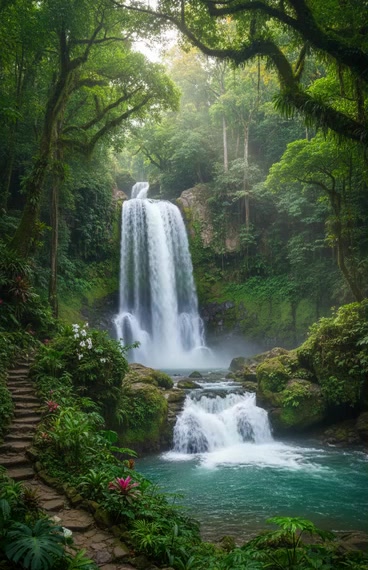} &
&
\includegraphics[width=0.15\textwidth]{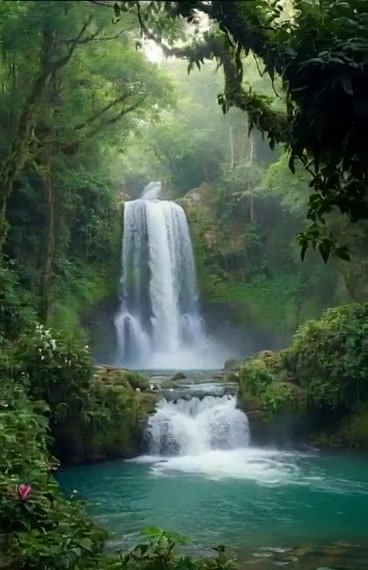} \\[1pt]
\rotatebox{90}{\small ~~Outpaint} &
\includegraphics[width=0.15\textwidth]{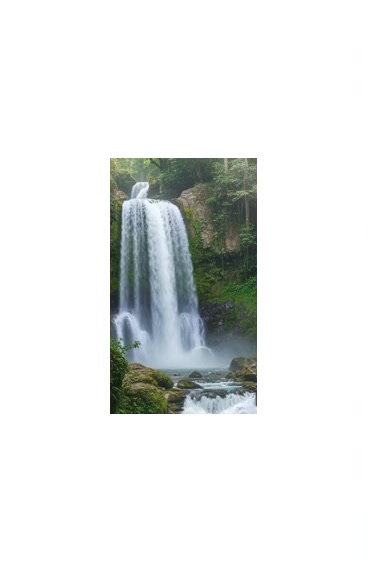} &
&
\includegraphics[width=0.15\textwidth]{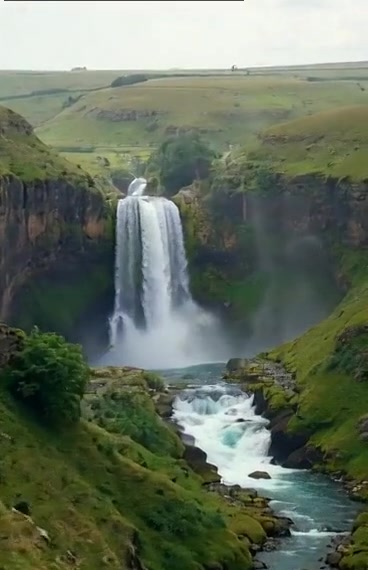} \\
\end{tabular}
\caption{Masked generation, layout control, and temporal extension. All outputs generated in real-time.}
\label{fig:qualitative}
\end{figure}

The adaptation supports structural control (V2V), masked generation including inpainting and outpainting (MV2V), temporal extension, reference-to-video (R2V, severely degraded; see Section~\ref{sec:limitations}), and arbitrary compositions of these modes. Dynamic masks can be driven by real-time object detectors such as YOLO~\citep{jocher2026yolo26}, and inpainting composes with LoRA~\citep{hu2022lora} for regional style transfer. Mode is inferred from provided inputs with no explicit mode parameter. Figure~\ref{fig:pipeline} shows the per-chunk processing flow; Figures~\ref{fig:structural} and~\ref{fig:qualitative} show representative outputs.

\section{Implementation}

The adaptation has been validated with the following Wan-based autoregressive pipelines:

\begin{table}[h]
\centering
\caption{Validated base pipelines.}
\label{tab:pipelines}
\begin{tabular}{ll}
\toprule
\textbf{Base Pipeline} & \textbf{Reference} \\
\midrule
LongLive & \citep{yang2026longlive} \\
StreamDiffusion V2 & \citep{feng2025streamdiffusionv2} \\
MemFlow & \citep{ji2025memflow} \\
Krea Realtime Video & \citep{millon2025krea} \\
Reward Forcing & \citep{lu2025rewardforcing} \\
\bottomrule
\end{tabular}
\end{table}

Hint projections are zero-initialized, enabling safe composition with LoRA and quantization. Mode is detected implicitly from provided inputs. In extension mode, reference hints are computed once and reused across subsequent chunks. We use the default VACE context scale ($\alpha = 1.0$) and did not tune it for this work.

\paragraph{Cache management for dual-stream encoding.}
VACE encodes conditioning frames into two streams: \emph{inactive} (preserved regions, masked to zero where generation occurs) and \emph{reactive} (generated regions, masked to zero where content is preserved). These are channel-concatenated before entering Context Blocks. In streaming contexts, the temporal autoencoder (TAE/3D VAE) maintains internal state across chunks. We allocate separate encoder caches for the inactive and reactive streams to prevent cross-contamination of temporal statistics. Cache behavior varies by mode, as summarized in Table~\ref{tab:cache}.

\begin{table}[h]
\centering
\caption{TAE encoder cache strategy per mode. Conditioning modes cache both streams (both carry temporally coherent signals). Inpainting/outpainting skip the reactive cache because reactive regions contain newly generated content each chunk; temporal blending from cached state introduces ghosting artifacts.}
\label{tab:cache}
\begin{tabular}{lcc}
\toprule
\textbf{Mode} & \textbf{Inactive Cache} & \textbf{Reactive Cache} \\
\midrule
Conditioning (depth, pose, etc.) & Enabled & Enabled \\
Inpainting / Outpainting & Enabled & Skipped \\
Extension (I2V) & Enabled & Skipped \\
\bottomrule
\end{tabular}
\end{table}

\section{Performance}

All models use bfloat16 precision with TAE decoder. Each configuration was run for 15 measured chunks after 3 warmup chunks. These are inference-only measurements; end-to-end throughput including streaming overhead is not measured here. Hardware and resolution details are noted per table.

To validate cross-model compatibility, we benchmark on two Wan2.1 model scales: LongLive (1.3B, 4 denoising steps, 12 frames per chunk) and Krea Realtime Video (14B). The same VACE adaptation code is used for both without modification.

\begin{table}[h]
\centering
\caption{LongLive 1.3B ablation (12 frames per chunk, 4 denoising steps, $368 \times 640$, NVIDIA RTX 5090 32\,GB, SageAttention~\citep{zhang2025sageattention}).}
\label{tab:perf-longlive}
\begin{tabular}{lcccc}
\toprule
\textbf{Configuration} & \textbf{Avg Latency} & \textbf{Avg FPS} & \textbf{Peak FPS} & \textbf{Peak VRAM} \\
\midrule
Baseline (no VACE) & 539\,ms & 22.3 & 22.8 & 13.2\,GB \\
+ Depth Control & 698\,ms & 17.2 & 17.4 & 14.6\,GB \\
+ Inpainting & 698\,ms & 17.2 & 17.3 & 14.6\,GB \\
+ Extension (I2V) & 534\,ms & 22.5 & 22.7 & 14.6\,GB \\
\bottomrule
\end{tabular}
\end{table}

\begin{table}[h]
\centering
\caption{Krea Realtime Video 14B ablation (12 frames per chunk, bfloat16, $320 \times 576$, NVIDIA H100 80\,GB). FlashAttention 2 used; SageAttention 2.2.0 produces artifacts on Hopper GPUs.}
\label{tab:perf-krea}
\begin{tabular}{lcccc}
\toprule
\textbf{Configuration} & \textbf{Avg Latency} & \textbf{Avg FPS} & \textbf{Peak FPS} & \textbf{Peak VRAM} \\
\midrule
Baseline (no VACE) & 741\,ms & 16.2 & 16.2 & 45.2\,GB \\
+ Depth Control & 887\,ms & 13.5 & 13.6 & 45.1\,GB \\
+ Inpainting & 958\,ms & 12.5 & 12.6 & 45.1\,GB \\
+ Extension (I2V) & 965\,ms & 12.4 & 12.5 & 45.1\,GB \\
\bottomrule
\end{tabular}
\end{table}

\begin{table}[h]
\centering
\caption{Control adherence metrics for VACE streaming controls (LongLive 1.3B, 5 chunks, $368 \times 640$). Depth RMSE: root mean squared error between input depth maps and depth extracted from the generated output via Video Depth Anything (lower is better). Mask Preservation: SSIM between input and output in unmasked regions (higher is better). We report metrics for the two control modes that admit straightforward objective measurement. These metrics confirm that controls function in streaming mode; absolute values are not directly comparable to batch-mode benchmarks such as ControlNet++~\citep{li2024controlnetpp}, which use different depth estimators and image scales.}
\label{tab:quality}
\begin{tabular}{lcc}
\toprule
\textbf{Configuration} & \textbf{Depth RMSE $\downarrow$} & \textbf{Mask Preservation $\uparrow$} \\
\midrule
+ Depth Control & 0.157 & --- \\
+ Inpainting & --- & 0.983 \\
\bottomrule
\end{tabular}
\end{table}

VACE adds approximately 1.4\,GB of VRAM overhead for the 1.3B model (the VACE Context Block weights). At the 14B scale, VRAM overhead is negligible relative to the base model's ${\sim}$45\,GB footprint. Per-chunk latency overhead is consistent across scales: depth control adds 20--30\% and inpainting adds 29--30\%. With LongLive, extension mode caches reference hints after the first chunk, resulting in negligible overhead (${\sim}$1\%) for subsequent chunks; this benefit is less pronounced at 14B where the base model dominates latency.

Note that the two benchmarks use different attention backends: SageAttention on the RTX 5090 (Table~\ref{tab:perf-longlive}) and FlashAttention 2 on the H100 (Table~\ref{tab:perf-krea}). The VACE overhead percentages are comparable across both configurations.

\section{Related Work}

The primary alternative for real-time controlled video generation is MotionStream~\citep{shin2025motionstream}. MotionStream achieves higher quality for its trajectory-based control modality through full-model distillation from a bidirectional teacher. In contrast, this adaptation reuses pretrained VACE weights without any additional training and supports multiple control types (depth, scribble, optical flow, layout, masking, and reference guidance) with arbitrary compositions. This flexibility comes at the cost of per-task quality relative to specialized distilled models.

\section{Limitations}
\label{sec:limitations}

\begin{itemize}
    \item \textbf{Temporal coherence} can degrade over extended generations ($100$+ frames) without re-anchoring, a general consequence of autoregressive generation.
    \item \textbf{Control signal variance:} Some signals (depth, scribble, layout) work reliably; others require more tuning.
    \item \textbf{Reference-to-Video} is the most problematic capability. Detail preservation and reference fidelity are severely degraded compared to batch VACE due to causal attention and per-chunk processing. Further architectural work is needed.
    \item \textbf{First+last frame extension} has reduced utility compared to batch VACE due to small chunk sizes in streaming contexts.
    \item \textbf{No perceptual quality comparison} to batch VACE is provided. The autoregressive base models (e.g., LongLive) are separately trained models that share Wan2.1's architecture but differ in attention pattern, step count, and training data. A comparison against batch Wan2.1 with VACE would conflate these base model differences with the adaptation itself, making it impossible to isolate the quality impact of our architectural change.
\end{itemize}

\section{Conclusion}

By moving reference frames from the diffusion latent space into a parallel conditioning pathway, this adaptation preserves the fixed chunk sizes and KV caching that autoregressive models require while reusing existing VACE weights directly. Structural control, masked generation, and temporal extension add 20--30\% latency overhead with negligible VRAM cost. The approach has been validated across Wan2.1 1.3B and 14B model scales without per-model modifications.

\section*{Acknowledgements}

The author thanks Yondon Fu, Rafal Leszko, and Marco Tundo for their support and feedback throughout this work.

\bibliographystyle{plainnat}
\bibliography{references}

@inproceedings{jiang2025vace,
  title={{VACE}: All-in-One Video Creation and Editing},
  author={Jiang, Zeyinzi and Han, Zhen and Mao, Chaojie and Zhang, Jingfeng and Pan, Yulin and Liu, Yu},
  booktitle={Proceedings of the IEEE/CVF International Conference on Computer Vision (ICCV)},
  year={2025},
  note={arXiv:2503.07598}
}

@inproceedings{yang2026longlive,
  title={{LongLive}: Real-time Interactive Long Video Generation},
  author={Yang, Shuai and Huang, Wei and Chu, Ruihang and Xiao, Yicheng and Zhao, Yuyang and Wang, Xianbang and Li, Muyang and Xie, Enze and Chen, Yingcong and Lu, Yao and Han, Song and Chen, Yukang},
  booktitle={International Conference on Learning Representations (ICLR)},
  year={2026},
  note={arXiv:2509.22622}
}

@misc{millon2025krea,
  title={Krea Realtime 14B: Real-time Video Generation},
  author={Millon, Erwann},
  year={2025},
  howpublished={\url{https://github.com/krea-ai/realtime-video}}
}

@article{feng2025streamdiffusionv2,
  title={{StreamDiffusionV2}: A Streaming System for Dynamic and Interactive Video Generation},
  author={Feng, Tianrui and Li, Zhi and Yang, Shuo and Xi, Haocheng and Li, Muyang and Li, Xiuyu and Zhang, Lvmin and Yang, Keting and Peng, Kelly and Han, Song and Agrawala, Maneesh and Keutzer, Kurt and Kodaira, Akio and Xu, Chenfeng},
  journal={arXiv preprint arXiv:2511.07399},
  year={2025}
}

@article{shin2025motionstream,
  title={{MotionStream}: Real-Time Video Generation with Interactive Motion Controls},
  author={Shin, Joonghyuk and Li, Zhengqi and Zhang, Richard and Zhu, Jun-Yan and Park, Jaesik and Shechtman, Eli and Huang, Xun},
  journal={arXiv preprint arXiv:2511.01266},
  year={2025}
}

@inproceedings{peebles2023dit,
  title={Scalable Diffusion Models with Transformers},
  author={Peebles, William and Xie, Saining},
  booktitle={Proceedings of the IEEE/CVF International Conference on Computer Vision (ICCV)},
  year={2023},
  note={arXiv:2212.09748}
}

@article{wang2025wan,
  title={Wan: Open and Advanced Large-Scale Video Generative Models},
  author={{Team Wan} and Wang, Ang and Ai, Baole and Wen, Bin and Mao, Chaojie and Xie, Chen-Wei and Chen, Di and Yu, Feiwu and Zhao, Haiming and Yang, Jianxiao and Zeng, Jianyuan and Wang, Jiayu and Zhang, Jingfeng and Zhou, Jingren and Wang, Jinkai and Chen, Jixuan and Zhu, Kai and Zhao, Kang and Yan, Keyu and Huang, Lianghua and Feng, Mengyang and Zhang, Ningyi and Li, Pandeng and Wu, Pingyu and Chu, Ruihang and Feng, Ruili and Zhang, Shiwei and Sun, Siyang and Fang, Tao and Wang, Tianxing and Gui, Tianyi and Weng, Tingyu and Shen, Tong and Lin, Wei and Wang, Wei and Zhou, Wenmeng and Wang, Wente and Shen, Wenting and Yu, Wenyuan and Shi, Xianzhong and Huang, Xiaoming and Xu, Xin and Kou, Yan and Lv, Yangyu and Li, Yifei and Liu, Yijing and Wang, Yiming and Zhang, Yingya and Huang, Yitong and Li, Yong and Wu, You and Liu, Yu and Pan, Yulin and Zheng, Yun and Hong, Yuntao and Shi, Yupeng and Feng, Yutong and Jiang, Zeyinzi and Han, Zhen and Wu, Zhi-Fan and Liu, Ziyu},
  journal={arXiv preprint arXiv:2503.20314},
  year={2025}
}

@misc{jocher2026yolo26,
  title={Ultralytics {YOLO26}},
  author={Jocher, Glenn and Qiu, Jing},
  year={2026},
  howpublished={\url{https://github.com/ultralytics/ultralytics}},
  note={Ultralytics v8.4.0}
}

@inproceedings{hu2022lora,
  title={{LoRA}: Low-Rank Adaptation of Large Language Models},
  author={Hu, Edward J. and Shen, Yelong and Wallis, Phillip and Allen-Zhu, Zeyuan and Li, Yuanzhi and Wang, Shean and Wang, Lu and Chen, Weizhu},
  booktitle={International Conference on Learning Representations (ICLR)},
  year={2022},
  note={arXiv:2106.09685}
}

@article{su2021roformer,
  title={{RoFormer}: Enhanced Transformer with Rotary Position Embedding},
  author={Su, Jianlin and Lu, Yu and Pan, Shengfeng and Murtadha, Ahmed and Wen, Bo and Liu, Yunfeng},
  journal={arXiv preprint arXiv:2104.09864},
  year={2021}
}

@inproceedings{zhang2025sageattention,
  title={{SageAttention}: Accurate 8-Bit Attention for Plug-and-play Inference Acceleration},
  author={Zhang, Jintao and Wei, Jia and Huang, Haofeng and Zhang, Pengle and Zhu, Jun and Chen, Jianfei},
  booktitle={International Conference on Learning Representations (ICLR)},
  year={2025},
  note={arXiv:2410.02367}
}

@article{ji2025memflow,
  title={{MemFlow}: Flowing Adaptive Memory for Consistent and Efficient Long Video Narratives},
  author={Ji, Sihui and Chen, Xi and Yang, Shuai and Tao, Xin and Wan, Pengfei and Zhao, Hengshuang},
  journal={arXiv preprint arXiv:2512.14699},
  year={2025}
}

@inproceedings{li2024controlnetpp,
  title={{ControlNet++}: Improving Conditional Controls with Efficient Consistency Feedback},
  author={Li, Ming and Yang, Taojiannan and Kuang, Huafeng and Wu, Jie and Wang, Zhaoning and Xiao, Xuefeng and Chen, Chen},
  booktitle={European Conference on Computer Vision (ECCV)},
  year={2024},
  note={arXiv:2404.07987}
}

@article{lu2025rewardforcing,
  title={Reward Forcing: Efficient Streaming Video Generation with Rewarded Distribution Matching Distillation},
  author={Lu, Yunhong and Zeng, Yanhong and Li, Haobo and Ouyang, Hao and Wang, Qiuyu and Cheng, Ka Leong and Zhu, Jiapeng and Cao, Hengyuan and Zhang, Zhipeng and Zhu, Xing and Shen, Yujun and Zhang, Min},
  journal={arXiv preprint arXiv:2512.04678},
  year={2025}
}

\end{document}